\documentclass[letterpaper]{article} % DO NOT CHANGE THIS
\usepackage{aaai25}  % DO NOT CHANGE THIS
\usepackage{times}  % DO NOT CHANGE THIS
\usepackage{helvet}  % DO NOT CHANGE THIS
\usepackage{courier}  % DO NOT CHANGE THIS
\usepackage[hyphens]{url}  % DO NOT CHANGE THIS
\usepackage{graphicx} % DO NOT CHANGE THIS
\usepackage{natbib}  % DO NOT CHANGE THIS AND DO NOT ADD ANY OPTIONS TO IT
\usepackage{caption} % DO NOT CHANGE THIS AND DO NOT ADD ANY OPTIONS TO IT
\usepackage{algorithm}
\usepackage{algorithmic}
\usepackage{amsmath}
\usepackage{amssymb}
\usepackage{amsthm}
\usepackage{subcaption}
\usepackage{multirow}
\newtheorem{assumption}{Assumption}
\newtheorem{lemma}{Lemma}
\newtheorem{theorem}{Theorem}

\usepackage{newfloat}
\usepackage{listings}
\DeclareCaptionStyle{ruled}{labelfont=normalfont,labelsep=colon,strut=off} % DO NOT CHANGE THIS
\floatstyle{ruled}
\newfloat{listing}{tb}{lst}{}
\floatname{listing}{Listing}
\title{Tilted Quantile Gradient Updates for Quantile-Constrained \\ Reinforcement Learning}
\author{
    Chenglin Li\textsuperscript{\rm 1},
    Guangchun Ruan\textsuperscript{\rm2},
    Hua Geng\textsuperscript{\rm 1}\thanks{Corresponding author.}\\
}
\affiliations{
    \textsuperscript{\rm 1}Department of Automation, Tsinghua University, Beijing, China\\
    \textsuperscript{\rm 2}Laboratory for Information \& Decision Systems, MIT, Boston, USA\\
    licl23@mails.tsinghua.edu.cn, gruan@mit.edu, genghua@tsinghua.edu.cn
}

\usepackage{bibentry}
\begin{document}

\maketitle

\begin{abstract}
Safe reinforcement learning (RL) is a popular and versatile paradigm to learn reward-maximizing policies with safety guarantees. Previous works tend to express the safety constraints in an expectation form due to the ease of implementation, but this turns out to be ineffective in maintaining safety constraints with high probability. To this end, we move to the quantile-constrained RL that enables a higher level of safety without any expectation-form approximations. We directly estimate the quantile gradients through sampling and provide the theoretical proofs of convergence. Then a tilted update strategy for quantile gradients is implemented to compensate the asymmetric distributional density, with a direct benefit of return performance. Experiments demonstrate that the proposed model fully meets safety requirements (quantile constraints) while outperforming the state-of-the-art benchmarks with higher return.
\end{abstract}


\section{Introduction}

There has been rising attention on Safe reinforcement learning (RL) which seeks to develop a policy that maximizes the expected cumulative return while meeting all the safety constraints. A common option is the Constrained Markov Decision Process (CMDP) framework, in which the safety constraints are established to bound the expected cumulative cost $C=\sum_t \gamma^t c$ by:
\begin{equation}
  \mathbb{E}[C]\le d
  \label{eq:ec}
\end{equation}

Most previous works follow this framework to construct constraints, because this form is consistent with the expected cumulative return (objective function) such that the gradient calculation would become the same for both the safety constraints and the objective function. As for the solution, the Lagrange approach is common~\cite{achiam2017constrained,liang2018accelerated, tessler2018reward, liu2020ipo}, but there are other options such as projection-based methods \cite{yang2020projection}, shielding methods \cite{alshiekh2018safe,carr2023safe}, and barrier methods \cite{marvi2021safe,cheng2019end}.

Unfortunately, the above expectation setup does not apply for many real-world safety-critical applications. Bounding expectations may still induce constraint violation in extreme cases, and the potential risk cannot be strictly limited. A better solution is to apply probability-related constraints, e.g. 95\%-quantile, to impose a more accurate and robust requirement for safety. In this case, the probability $\Pr[C\le d]$ can be more informative than the safety expectation $\mathbb{E}[C]\le d$.

A probabilistic constraint or a chance constraint is typically expressed as follows~\cite{chow2018risk,chen2024probabilistic}:
\begin{equation}
  \Pr[C\le d]\ge 1-\varepsilon
  \label{eq:cc}
\end{equation}

Eqn.~(\ref{eq:cc}) strictly limits the probability of constraint violation under a given level $\varepsilon$, but this constraint is computationally intractable and suffers from low sample efficiency and lack of distribution priors. A direct transformation of (\ref{eq:cc}) needs to apply the quantile, or value-at-risk (VaR) metric. Mathematically, Eqn.~(\ref{eq:cc}) is equivalent to the following quantile constraint:
\begin{equation}
  q_{1-\varepsilon}:=\inf\{x|\Pr[C\le x]\ge 1-\varepsilon\}\le d
  \label{eq:qc}
\end{equation}

An optimization model with quantile constraints integrated is still computationally intensive for training. In the literature, quantile optimization has been widely studied. Estimating the gradient of quantile is a challenge, and most of the existing solutions such as the perturbation analysis~\cite{jiang2015estimating}, the likelihood ratio method~\cite{glynn2021computing}, and the kernel density estimation~\cite{hong2009simulating} are heavily relied on the analytical model formulation. Also, these models are focused on unconstrained optimization problems with quantile-based objectives, but the focus of this paper is on the quantile constrains for Safe RL.

Quantile-constrained RL was first studied in \cite{jung2022quantile}, and the idea was to supplement the expected cumulative cost $\mathbb{E}[C]$ with an additive term to approximate the quantile $q_{1-\varepsilon}$. Within this setting, the quantile constraint could be converted into an expectation-type constraint at last, aligning with the CMDP framework. Note that this work required the cumulative cost distribution, and its empirical performances might be over-conservative (with relatively low return) because of the biased approximation of quantiles.

Other related works have also been conducted to bound the probability of constraint violation. Yang et al. (2023) applied Conditional Value-at-Risk (CVaR) as an approximation of the quantile. Chow et al. (2018) proposed a trajectory-based method with chance constraints to bound the probability of constraint violation. However, this trajectory-based approach updates the policy only once based on a batch of trajectories, resulting in low sample efficiency, which is not suitable for practical application. Some model-based methods have also been proposed to guarantee the safety probability \cite{peng2022model, pfrommer2022safe}, but these method requires prior knowledge of the environment, which is not practical in many real-world scenarios.

In this paper, we establish a novel quantile-constrained RL model, namely Tilted Quantile Policy Optimization (TQPO) model, where the safety constraints are expressed in a quantile form as Eqn.~(\ref{eq:qc}). For the estimation of quantile gradients, we get rid of any expectation-form approximations and directly estimate the quantile gradient through a sampling technique. To avoid over-conservatism of the policy and gain higher return, a tilted quantile gradient update is designed to compensate the asymmetric distributional density of quantiles.

The rest of this paper is structured as follows. We first introduce the background in the Preliminaries section. Then we present the methodology of the TQPO model, followed by a convergence analysis. In the Experiments section, simulation results demonstrate that the proposed model fully guarantee safety while outperforming the state-of-the-art benchmarks with higher returns.

\section{Preliminaries}

\subsection{Constrained Markov Decision Process}\label{sec:cmdp}

A CMDP framework~\cite{altman2021constrained} is characterized by a tuple $\langle S,A,P,r,c,d,\gamma \rangle$, where $S$ denotes the state space, $A$ denotes the action space, $P(\cdot|s,a)$ is the state transition probability function, $r(s,a)$ is the reward function, $c(s,a)$ is the cost function, $d$ is a given threshold, and $\gamma\in(0,1)$ is the discount factor.

The agent interacts with the environment at each time step $t$ by observing the current state $s_t\in S$ and selecting an action $a_t\in A$, then receives a reward $r(s_t,a_t)$ as well as a cost $c(s_t,a_t)$. This agent focuses on learning a policy $\pi_\theta(\cdot|s)$ parameterized by $\theta$. The next states are generated by the state transition probability function $P(\cdot|s,a)$ and the policy $\pi_\theta(\cdot|s)$. Given an initial state $s_0$, the cumulative return is defined as $R(s_0, \pi_\theta)=\sum_{t=0}^{\infty}\gamma^tr(s_t,a_t)$, and the cumulative cost is defined as $C(s_0, \pi_\theta)=\sum_{t=0}^{\infty}\gamma^tc(s_t,a_t)$.

Based on the previous definitions, CMDP framework is established to find a policy $\pi_\theta$ that maximizes the expected return while satisfying the expected cost under a given threshold $d$:
\begin{align}
\max_{\theta}& \ V(s, \pi_\theta):=\mathbb{E}_{\pi_\theta}[R(s, \pi_\theta)]=\mathbb{E}_{\pi_\theta}[\sum_{t}\gamma^tr(s_t,a_t)] \notag \\
\text{s.t.}& \ \mathbb{E}_{\pi_\theta}[C(s, \pi_\theta)]=\mathbb{E}_{\pi_\theta}[\sum_{t}\gamma^tc(s_t,a_t)]\le d
\label{eq:ecrl}
\end{align}
where $V(s, \pi_\theta)$ denotes the state value function. The expectation term of constraints matches the same term of the objective, therefore the gradient of the constraint can be calculated in a similar way, resulting in easy implementation by the Lagrangian method. However, the expectation-based formulation of the constraints cannot strictly limit the probability of constraint violation, which is not suitable for safety-critical applications.

\subsection{Quantile-constrained RL}

For a sequence of samples $\{s_0, s_1,\ldots, s_N\}$, the cumulative cost $C(s_i, \pi_\theta)$ of each state $s_i$ follows the empirical distribution $F(\cdot; \pi_\theta)$. Given a probability level $\varepsilon\in(0,1)$, similar to Eqn.~(\ref{eq:qc}) with the cumulative cost $C(s, \pi_\theta)$ following the distribution $F(\cdot; \pi_\theta)$, the $1-\varepsilon$ quantile of $C(s, \pi_\theta)$ can be rewritten as:
\begin{equation}
  q_{1-\varepsilon}(\pi_\theta):=\inf\{q|\Pr(C(s, \pi_\theta)\le q)=F(q; \pi_\theta)\ge 1-\varepsilon\}
  \label{eq:quantile}
\end{equation}

When the quantile of the distribution of $C$ satisfies $q_{1-\varepsilon}(\pi_\theta) \le d$, the probability of constraint violation is under $\varepsilon$. Therefore, the quantile-constrained RL problem can be formulated as follows:
\begin{align}
    \max_{\theta}& \ V(s, \pi_\theta)=\mathbb{E}_{\pi_\theta}[R(s, \pi_\theta)]=\mathbb{E}_{\pi_\theta}[\sum_{t}\gamma^tr(s_t,a_t)] \notag \\
    \text{s.t.}& \ q_{1-\varepsilon}(\pi_\theta)\le d
    \label{eq:qcrl}
    \end{align}
    
Since the quantile $q_{1-\varepsilon}(\pi_\theta)$ is not of an expectation form, the conventional expectation Bellman equation cannot solve the gradient calculation of the quantile constraint. How to estimate quantile gradients become a challenge in this problem setting.

\section{Methodology}
\label{sec:method}

In this section, we use a sample-based approach to estimate the gradient of the quantile constraint, and then construct a tilted quantile gradient update to accelerate the training process. Finally, we proposed a quantile-constrained RL algorithm based on the tilted quantile gradient update.

\subsection{Estimating Quantile Gradients Through Sampling}

Consider a quantile constraint as Eqn.~(\ref{eq:quantile}). When $F(\cdot; \pi_\theta)$ is continuous and differentiable (a minor assumption), the quantile $q_{1-\varepsilon}(\pi_\theta)$ is mathematically the inverse of $F(\cdot; \pi_\theta)$. According to the inverse function theorem, the gradient of this quantile constraint can be calculated as follows:
\begin{equation}
  \fontsize{9.5}{8}\selectfont
    \nabla_\theta q_{1-\varepsilon}(\pi_\theta)=\nabla_\theta F^{-1}(1-\varepsilon; \pi_\theta)=-\frac{\nabla_\theta F(q; \pi_\theta)}{f(q; \pi_\theta)}\Big|_{q=q_{1-\varepsilon}(\pi_\theta)}
  \label{eq:qgrad}
\end{equation}
where $f(\cdot; \pi_\theta)$ is the probability density function (PDF) of the cumulative cost $C$. Eqn.~(\ref{eq:qgrad}) implies that the quantile gradient can be estimated by figuring out the above numerator and denominator.

The numerator $\nabla_\theta F(q; \pi_\theta)$ can be estimated by the likelihood ratio method in policy gradient algorithms. Given a batch of samples $\{s_0, s_1,... s_N\}$, the gradient of the CDF can be estimated as follows:
\begin{equation}
  \fontsize{8.5}{7}\selectfont
  \begin{aligned}
    \nabla_\theta F(q; \pi_\theta)=&\nabla_\theta \mathbb{E}[I(C(s, \pi_\theta)\ge q)]=- \nabla_\theta \mathbb{E}[I(C(s, \pi_\theta)\le q)]\\
    % =&\nabla_\theta \mathbb{E}[1-I(C(s, \pi_\theta)\le q)]\\
    % =&- \nabla_\theta \mathbb{E}[I(C(s, \pi_\theta)\le q)]\\
    =&- \mathbb{E}[I(C(s, \pi_\theta)\le q)\sum_{t=i}^{N-1}\nabla_\theta\log\pi_\theta(a_t|s_t)]\\
    \approx& -\frac{1}{N}\sum_{i=1}^{N}I(C(s_i, \pi_\theta)\le q)\sum_{t=i}^{N-1}\nabla_\theta\log\pi_\theta(a_t|s_t)
  \label{eq:gradcdf}
  \end{aligned}
\end{equation}
where $q=q_{1-\varepsilon}(\pi_\theta)$ denotes the $1-\varepsilon$ quantile, which is unknown in practice. We adopt a iterative method to estimate the quantile $q_{1-\varepsilon}(\pi_\theta)$ with $q_k$ from every batch of samples as follows:
\begin{equation}
    q_{k+1}=q_k+\alpha(\hat q_{1-\varepsilon}-q_k)
  \label{eq:quantileupdate}
\end{equation}
where $\hat q_{1-\varepsilon}$ is the empirical quantile of the cumulative cost $C$ from the batch of samples, and $\alpha\in(0,1)$ is the update rate for smoothness. 

The denominator $f(q; \pi_\theta)$ in Eqn.~(\ref{eq:qgrad}) is the PDF of the cumulative cost $C$, which is difficult to estimate without the priori of the environment \cite{jiang2022quantile}. Since PDF is always positive, the gradient of quantile in Eqn.~(\ref{eq:qgrad}) is in the same direction as $-\nabla_\theta F(q; \pi_\theta) |_{q=q_{1-\varepsilon}(\pi_\theta)}$, which can be adopted as an approximation of the gradient of the quantile constraint.

Collectively, the quantile gradient $\nabla_\theta q_{1-\varepsilon}(\pi_\theta)$ can be estimated by applying Eqn.~(\ref{eq:qgrad})--(\ref{eq:quantileupdate}). Therefore, policy gradient method is applicable solve the quantile-constrained RL problem in Eqn.~(\ref{eq:qcrl}). Applying the Lagrangian method, the dual objective of the quantile-constrained RL problem is defined as follows:
\begin{equation}
  \min_{\lambda\ge 0}\max_{\theta}\mathcal{L}(\theta, \lambda, q)=V(s, \pi_\theta)-\lambda(q_{1-\varepsilon}(\pi_\theta)-d)
\label{eq1:loss}
\end{equation}
Given a batch of samples $\{s_0, s_1,\ldots, s_N\}$, the gradient of the dual objective in Eqn.~(\ref{eq1:loss}) w.r.t. policy parameter $\theta$ can be calculated as follows. This gradient can then be used to update the policy parameter $\theta$:
\begin{equation}
  \fontsize{8.5}{9}\selectfont
  \begin{aligned}
    &\nabla_\theta \mathcal{L}(\theta, \lambda, q)=\nabla_\theta V(s, \pi_\theta)-\lambda\nabla_\theta q_{1-\varepsilon}(\pi_\theta)\\
    \approx&\frac{1}{N}\sum_{i=1}^{N}\Big [V(s_i, \pi_\theta)-\lambda I(C(s_i, \pi_\theta)\le q_k)\Big ]\sum_{t=i}^{N-1}\nabla_\theta\log\pi_\theta(a_t|s_t)
  \label{eq1:grad}
  \end{aligned}
\end{equation}

For the Lagrangian multiplier $\lambda$, it can be updated by the gradient of the dual objective w.r.t. $\lambda$ as follows:
\begin{equation}
  \nabla_\lambda \mathcal{L}(\theta, \lambda, q)=-(q_{1-\varepsilon}(\pi_\theta)-d)
  \label{eq1:grad_lambda}
\end{equation}
Then the Lagrangian multiplier $\lambda$ can be updated as follows:
\begin{equation}
  \lambda\leftarrow \max\{\lambda+\eta (q_{1-\varepsilon}(\pi_\theta)-d), 0\}
  \label{eq1:lambda_update}
\end{equation}
where $\eta$ is the update rate of $\lambda$. 

\subsection{Tilted Quantile Gradient Update}

However, a direct use of Eqn.~(\ref{eq1:lambda_update}) with a fixed $\eta$ to update $\lambda$ can be inefficient due to the overshooting of $\lambda$ at the early stage of training. The reason is that the quantile $q_{1-\varepsilon}(\pi_\theta)$ may have an asymmetric distributional density around the threshold $d$. We illustrate this issue as follows.

Assuming a given violation probability level $\varepsilon$, at the early stage of training, the initial policy may behave unsafe with violation probability much larger than $\varepsilon$, as well as $q_{1-\varepsilon}(\pi_\theta)$ several times larger than the threshold $d$, resulting in a large increase of $\lambda$. This large increase makes $\lambda$ overshoot to a large value at the early training, which may result in over-conservatism of the policy \cite{peng2022model}. 

Later when the policy satisfies the constraint with $0 \le q_{1-\varepsilon}(\pi_\theta)\le d$, $\lambda$ starts to decrease with a slower rate. Even for an absolute safe policy with $q_{1-\varepsilon}(\pi_\theta)=0$ where $\lambda$ decreases fastest, the decrease of $\lambda$ is still slow as $\Delta\lambda=\eta(q_{1-\varepsilon}(\pi_\theta)-d)= -\eta  d$, rather than the rapid increase earlier with $q_{1-\varepsilon}(\pi_\theta)$ several times larger than $d$. The slow decrease of $\lambda$ from a large value may result in a slow recovery of the policy from over-conservatism. In general, this asymmetric distributional density of the quantile indicates that $q_{1-\varepsilon}(\pi_\theta)-d$ is relatively large in the early stage of training, but not small enough at the later stage. This issue makes $\lambda$ overshoot rapidly at the early unsafe training, while decrease slowly at the later over-conservatism stage, which may eventually result in slow convergence of the algorithm.

To address this issue, we propose a tilted quantile gradient update to compensate the asymmetric distributional density of $q_{1-\varepsilon}(\pi_\theta)$. Since we expect the distribution of $q_{1-\varepsilon}(\pi_\theta)$ to be symmetric around $d$, a tilted factor is designed to compensate the asymmetric distribution. Similar to the pinball loss in quantile regression \cite{steinwart2011estimating}, we revise the update rate $\eta$ in Eqn.~(\ref{eq1:lambda_update}) with a tilted term defined as follows:
\begin{equation}
  \eta= \left\{
    \begin{array}{ll}
      \eta_+=\frac{F_q(d)+\delta}{1+\delta}, & \text{if } q_{1-\varepsilon}(\pi_\theta)\ge d\\
      \eta_-=\frac{1-F_q(d)+\delta}{1+\delta}, & \text{if } q_{1-\varepsilon}(\pi_\theta)<d 
    \end{array}
  \right.
  \label{eq1:tilted}
\end{equation}
where $F_q(d)$ denotes the CDF of the distribution of quantile $q_{1-\varepsilon}(\pi_\theta)$ at $d$, which is estimated by sampling per epoch. $\eta_+$ and $\eta_-$ are the update rates for the positive and negative tilted terms respectively, $\delta\in(0,1)$ a small smoothing factor.

The tilted term in Eqn.~(\ref{eq1:tilted}) is utilized to update the Lagrangian multiplier $\lambda$ in Eqn.~(\ref{eq1:lambda_update}). For example, in the early stage of training, the policy is always unsafe with $q_{1-\varepsilon}(\pi_\theta)\ge d$, resulting in a small positive update rate $\eta=\eta_+\approx\delta$. With the increase of $\lambda$, the policy gradually satisfies the constraint with $q_{1-\varepsilon}(\pi_\theta)\le d$, then $\lambda$ switches to decrease with a negative update rate $\eta=\eta_-\approx1-\delta$. Assuming $\delta=0.1$, the decrease update rate $\eta_-\approx0.9$ will be about 9 times larger than the increase rate $\eta_+\approx0.1$. Therefore, 
the decrease of $\lambda$ from a large value can be accelerated, with the tilted term compensating the asymmetric distributional density of $q_{1-\varepsilon}(\pi_\theta)$, which facilitates the recovery of the policy from over-conservatism. 

The tilted term in Eqn.~(\ref{eq1:tilted}) is performed each epoch to update $\lambda$ adaptively, boosts the decrease of $\lambda$ from overshoot and tunes it to a more appropriate value range eventually, which can prevent the policy from over-conservatism and facilitate it to achieve higher return.

\subsection{Tilted Quantile Policy Optimization}
Based on the quantile gradient estimation with sampling and the tilted quantile gradient update, we propose an algorithm named Tilted Quantile Policy Optimization (TQPO) to solve the quantile-constrained RL problem in Eqn.~(\ref{eq:qcrl}). The algorithm is based on the classic RL algorithm Proximal Policy Optimization (PPO) \cite{schulman2017proximal} with the quantile constraint. We use a policy network parameterized by $\theta$ to represent the policy $\pi_\theta(\cdot|s)$, and a value network parameterized by $\phi$ to obtain the estimated value function $V_\phi(s)$.

The loss function of $\theta$ is defined as follows to train the policy network:
\begin{equation}
  \begin{aligned}
    L_\theta=-\mathbb{E}_{\pi}\Big [\min\Big (\frac{\pi(a_t|s_t)}{\pi_{\text{old}}(a_t|s_t)}A,\text{clip}(\frac{\pi(a_t|s_t)}{\pi_{\text{old}}(a_t|s_t)},\\
    1-r_{clip},1+r_{clip})A\Big )\Big ]
  \label{eq:loss}
  \end{aligned}
\end{equation}
where 
\begin{equation}
  \begin{aligned}
  A=r(s_i,a_i)+\gamma V_\phi(s_{i+1}, \pi_\theta)-V_\phi(s_i, \pi_\theta)\\
  -\lambda I(C(s_i, \pi_\theta)\le q_k)
  \end{aligned}
  \label{eq:advantage}
\end{equation}
Eqn.~(\ref{eq:advantage}) is the reward advantage function in PPO style, with the addition of the quantile constraint gradient. Additionally, we adopt the importance sampling technique and clipped surrogate objective in PPO, shown in Eqn.~(\ref{eq:loss}), to stabilize the training and improve sample efficiency.

Overall, the training process of TQPO iterates as follows:
\begin{itemize}
  \item Generate a batch of samples $\{s_0, s_1,\ldots, s_N\}\sim\pi_\theta$
  \item Update the value network parameter $\phi$
  \item Update three main parameters $q$, $\theta$ and $\lambda$ as follows:
  \begin{subequations}
    \begin{align}
        q_{k+1} &= q_k + \alpha_k (\hat{q}_{1-\varepsilon} - q_k) \label{eq:quantileupdaterevised} \\
        \theta_{k+1} &= \theta_k + \beta_k \nabla_\theta L_\theta \label{eq:policyupdate} \\
        \lambda_{k+1} &= \lambda_k + \eta_k (q_k - d) \label{eq:lagrangeupdate}
    \end{align}
    \label{eq:updates}
  \end{subequations}
\end{itemize}
where $\alpha_k$, $\beta_k$, and $\eta_k$ are the update rates of the three parameters respectively. Implementation details of TQPO can be found in Appendix B.

\section{Convergence Analysis}\label{sec:convergence}

In this section, we provide the theoretical proofs of the convergence of the TQPO algorithm. First, let's reconsider the update of the three main parameters in the TQPO algorithm, i.e., the estimated quantile $q$, the policy parameter $\theta$ and the Lagrange multiplier $\lambda$ in Eqn.~(\ref{eq:updates}). For the convenience of theoretical analysis, We modify Eqn.~(\ref{eq:policyupdate}) by replacing the implemented loss $L_\theta$ to the Lagrange objective function $\mathcal{L}(\theta,\lambda,q)$ in Eqn.~(\ref{eq1:loss}). In order to prove the convergence, we first adopt the following assumptions:
% type assumption 1 here
\begin{assumption}\label{assumption1} For any probability level $\varepsilon\in(0,1)$, the objective function $\mathcal{L}(\theta,\lambda,q)$ is continuous and differentiable with respect to $\theta$.
\end{assumption}

\begin{assumption}\label{assumption2} $\nabla_\theta \mathcal{L}(\theta,\lambda,q)$ is Lipschitz continuous w.r.t. $\theta$, $\lambda$ and $q$, i.e., $\forall (\theta_1,\lambda_2,q_2), (\theta_2,\lambda_2,q_2)\in \Theta\times \mathbb{R_+}\times \mathbb{R}$, there exists a constant $\kappa$ such that $\|\nabla_\theta \mathcal{L}(\theta_1,\lambda_1,q_1)-\nabla_\theta \mathcal{L}(\theta_2,\lambda_2,q_2)\|\le \kappa\|(\theta_1,\lambda_1,q_1)-(\theta_2,\lambda_2,q_2)\|$, where $\Theta$ is the parameter space of the policy network.
\end{assumption}

\begin{assumption}\label{assumption3} The update rates $\alpha_k$, $\beta_k$, and $\eta_k$ are all positive, nonsummable, and square summable. In specific, this indicates that $\alpha_k>0, \sum_{k=0}^{\infty}\alpha_k=\infty,\sum_{k=0}^{\infty}\alpha_k^2<\infty$. Moreover, the update rates satisfy: $\eta_k=o(\beta_k), \beta_k=o(\alpha_k)$.
\end{assumption}

Assumption \ref{assumption1} is a common condition in continuous optimization, which ensures the continuity of the objective function w.r.t the policy parameter $\theta$. Assumption \ref{assumption2} and \ref{assumption3} are standard conditions for stochastic approximation analysis \cite{borkar2008stochastic, gattami2021reinforcement}. 

Assumption \ref{assumption3} indicates the update timescales of the quantile $q$, the policy parameter $\theta$, and the Lagrange multiplier $\lambda$, respectively, where $q$ is updated fastest, followed by $\theta$, and $\lambda$ is the slowest. The three parameters affect each other in the updating but with different timescales. Therefore, we can utilize the timescale separation to conduct the proof by two steps, first proving the convergence of $(\theta, q)$, and then proving the convergence of $(\theta, q, \lambda)$.

\subsection{Convergence of $(\theta, q)$}
First, we consider the convergence of the quantile $q$ and the policy parameter $\theta$ in the TQPO algorithm. Since $\lambda$ updates slower than $q$ and $\theta$, we can regard $\lambda$ as an arbitrary constant in the timescale of $q$ and $\theta$.

Considering the updates of $q$ and $\theta$ in Eqn.~(\ref{eq:quantileupdaterevised}) and Eqn.~(\ref{eq:policyupdate}), the two recursions are expected to track two coupled odinary differential equations (ODEs) with respect to $q$ and $\theta$:
\begin{equation}
  \begin{aligned}
    \dot q(t)&=g_1(\theta,q):=\hat q_{1-\varepsilon}(\theta)-q\\
    \dot \theta(t)&=g_2(\theta,q):=\nabla_\theta \mathcal{L}(\theta, q)
  \end{aligned}
  \label{eq:odeq}
\end{equation}

Since the update rate of $\theta$ is slower than $q$, we can regard $\theta$ as a constant when updating $q$.

\begin{lemma}
  For any $\overline{\theta}\in \Theta$, the ODE $\dot{q}(t)=g_1(\overline{\theta},q)$ has the unique global asymptotically stable equilibrium $q_{\overline{\theta}}$.
  \label{lemma1}
\end{lemma}

With the support of Lemma \ref{lemma1}, $\{q_k\}$ converges to the equilibrium of the ODE $\dot{q}(t)=g_1(\theta,q)$ for any $\theta\in \Theta$. Then we focus on the convergence of $\theta$. The gradient update of $\theta$ in Eqn.~(\ref{eq:policyupdate}) can be considered as tracking the right-hand side of the ODE in Eqn.~(\ref{eq:odeq}). Therefore, we adopt the following theorem to prove $\{\theta_k\}$ converge to the unique global asymptotically stable equilibrium.

\begin{theorem}
  For the two coupled iterations:\cite{borkar1997stochastic}
  \begin{equation}
    \begin{aligned}
      q_{k+1}=q_k+\alpha_k(g_1(\theta_k,q_k)+m_k)\\
      \theta_{k+1}=\theta+\beta_k(g_2(\theta_k,q_k)+n_k)
    \end{aligned}
    \label{eq:odeq_th} 
  \end{equation}
  for $k\ge 0$, where,
  \begin{itemize}
    \item(i): $g_1(\theta,q)$ and $g_2(\theta,q)$ are Lipschitz continuous
    \item(ii): $\alpha_k$ and $\beta_k$ satisfy Assumption \ref{assumption3}
    \item (iii): $m_k$ and $n_k$ are noise sequences and satisfy $\sum_{k=0}^{\infty}\alpha_k m_k,\ \ \sum_{k=0}^{\infty}\beta_k n_k<\infty$
  \end{itemize}

  If $\forall \overline{\theta}\in \Theta$, ODE $\dot{q}(t)=g_1(\overline{\theta},q)$ has a unique global asymptotically stable equilibrium point $q_{\overline{\theta}}$, the iterations \ref{eq:odeq_th} converge to the unique global asymptotically stable equilibrium of the ODE $\dot{\theta}(t)=g_2(\theta,q)$ a.s. on the set $\sup_k{|q_k|}<\infty$.
  \label{theorem1}
\end{theorem}

Theorem \ref{theorem1} requires $\{q_k\}$ and the log gradient of $\pi_\theta$ to be bounded, which can be guaranteed by the following lemma and assumption respectively.

\begin{lemma}
  If Assumptions \ref{assumption1}, \ref{assumption3} hold, the sequence $\{q_k\}$ satisfies $\sup_k|q_k|<\infty$.
  \label{lemma2}
\end{lemma}

\begin{assumption}
  The log gradient of the policy network $\nabla_\theta\log\pi(a|s,\theta)$ is bounded on the state space $S$ w.r.t. the policy parameter $\theta\in \Theta$, i.e., $\sup_{s\in S}\|\nabla_\theta\log\pi(a|s,\theta)\|<\infty$.
  \label{assumption4}
\end{assumption}

With the support of Lemma \ref{lemma2}, Theorem \ref{theorem1} indicates the sequence $\{\theta_k\}$ converges to the unique global asymptotically stable equilibrium of the ODE $\dot{\theta}(t)=g_2(\theta,q)$. So far, we have proved $(\theta,q)$ converge to their unique global asymptotically stable equilibriums. We then prove this converged $\theta$ is the optimal policy parameter with  following lemma:

\begin{lemma}
  If $\mathcal{L}(\theta,q)$ is strictly concave on $\Theta$, the ODE $\dot{\theta}(t)=g_2(\theta,q)$ has a unique global asymptotically stable equilibrium point $\theta^*=arg\max_{\theta}\mathcal{L}(\theta,q)$.
  \label{lemma3}
\end{lemma}

Lemma \ref{lemma3} indicate that the optimal policy parameter $\theta^*$ is the unique global asymptotically stable equilibrium point of the ODE $\dot{\theta}(t)=g_2(\theta,q)$. 

Above all, we first utilize Lemma \ref{lemma1} and Theorem \ref{theorem1} to prove the convergence of $(\theta,q)$ to the unique global asymptotically stable equilibriums of ODEs(\ref{eq:odeq}). Then we use Lemma \ref{lemma3} to prove the optimality of the converged $\theta$. Therefore, $(\theta,q)$ in TQPO algorithm converges to the optimal $(\theta^*,q^*)$ for a fixed $\lambda$. Next we provide the convergence analysis of the Lagrange multiplier $\lambda$.

\subsection{Convergence of $\lambda$}

Numerous works have proved the convergence of the Lagrange multiplier in two timescales constrained MDPs, where $(\theta, \lambda)$ converges to the optimal solution under certain conditions \cite{paternain2019constrained, gattami2021reinforcement}. As mentioned before, the update rate of $\lambda$ is slower than both $q$ and $\theta$, it is reasonable to merge the two faster parameters $\theta$ and $q$ into a new parameter $\theta '=(\theta , q)$. In the update process of $\lambda$, we can regard $\theta '$ converged to $\theta '^*=(\theta^*(\lambda) , q^*(\lambda))$. Therefore, the three timescales update of $(q,\theta,\lambda)$ can be considered as a two timescales update of $(\theta ',\lambda)$, and the standard analysis for constrained MDPs can be applied to the TQPO algorithm.

\begin{theorem}
  Under Assumptions \ref{assumption1}, \ref{assumption2}, \ref{assumption3}, \ref{assumption4}, if Slater's condition holds, the iterates $(\theta'_k, \lambda_k)=(q_k, \theta_k, \lambda_k)$ converge to the optimal solution a.s. \cite{borkar2008stochastic}
  \label{theorem2}
\end{theorem}

Theorem \ref{theorem2} is a standard analysis for the convergence of the dual problem in constrained MDPs in the RL literature. By Theorem \ref{theorem2}, we can conclude that the TQPO algorithm converges to the optimal solution of the quantile-constrained RL problem. Detailed proofs for the lemmas and theorems can be found in the Appendix A.

\section{Experiments}\label{sec:experiments}

\subsection{Simulation Setup}

We evaluate the proposed TQPO on three classic safe RL tasks: SimpleEnv, DynamicEnv and GremlinEnv from Mujoco and Safety Gym \cite{todorov2012mujoco, Ray2019}. 
\begin{figure}[h]
  \centering
  \subfloat[SimpleEnv]{\includegraphics[width=0.15\textwidth]{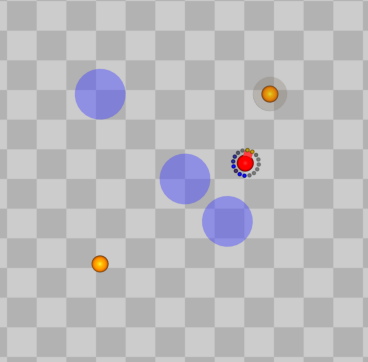}\label{fig:simp_env}}
  \hfill
  \subfloat[DynamicEnv]{\includegraphics[width=0.15\textwidth]{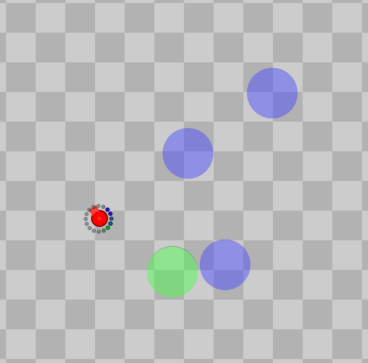}\label{fig:dyna_env}}
  \hfill
  \subfloat[GremlinEnv]{\includegraphics[width=0.15\textwidth]{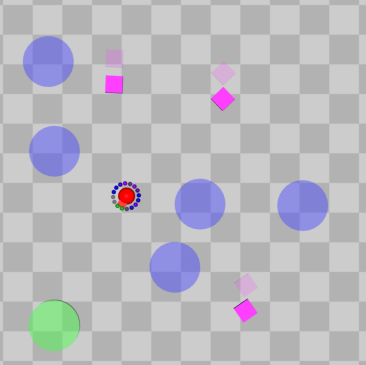}\label{fig:gem_env}}
  \caption{Safety Gym simulation environments}
  \label{fig:environments}
\end{figure}

In these tasks, a robot (red) is required to reach a goal while avoiding collisions with obstacles. The complexity of the three tasks gradually increases due to the addition of randomness. In SimpleEnv (Fig.~\ref{fig:simp_env}), the obstacles include fixed hazards (blue) and none-goal buttons (orange). When the robot reach the goal (orange covered by grey shadow), the environment swaps the goal and the none-goal button, therefore the new goal is generated deterministically. In DynamicEnv (Fig.~\ref{fig:dyna_env}), when the robot reaches the goal (green), a new goal is generated randomly. In GremlinEnv (Fig.~\ref{fig:gem_env}), the obstacles include moving gremlins (pink) and the goal is generated randomly.

\begin{figure*}[h]
  \centering
  \includegraphics[width=\textwidth]{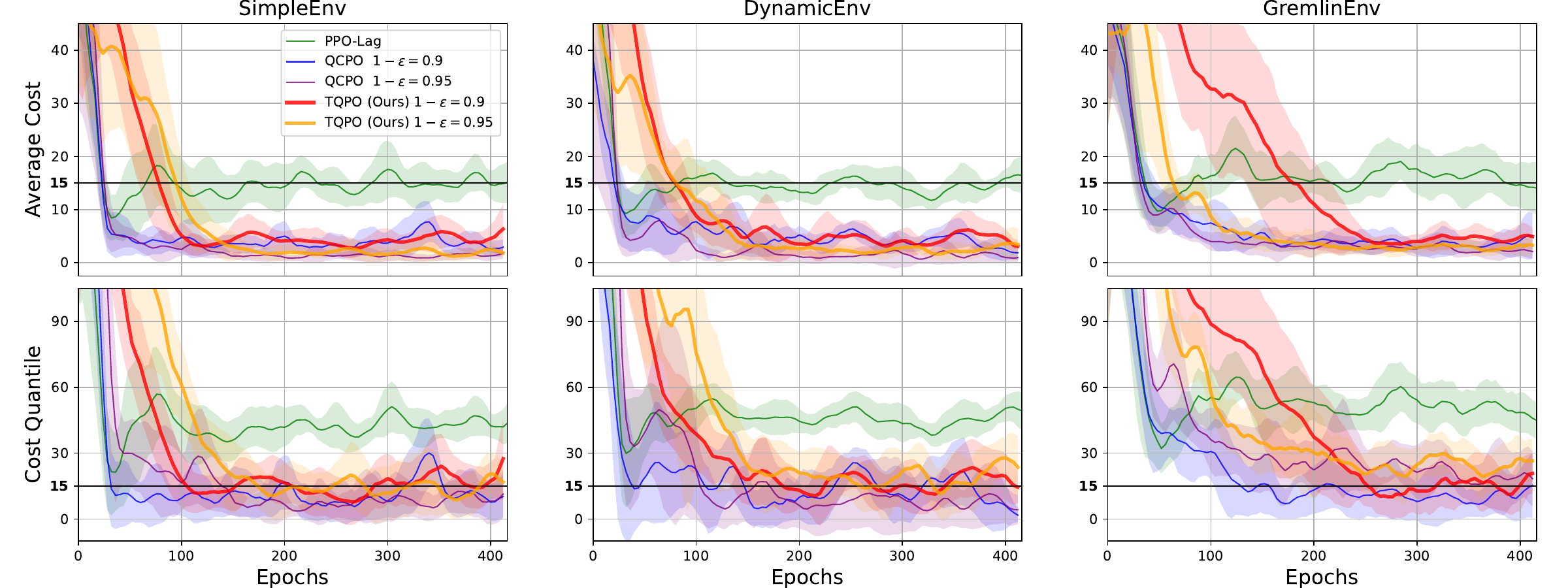}
  \caption{Average Cost (Row 1) and Cost Quantile (Row 2) of three algorithms on SimpleEnv (Column 1), DynamicEnv (Column 2) and GremlinEnv (Column 3). The cost quantile of PPO-Lag is calculated with $1-\varepsilon=90\%$}
  \label{fig:results_cost}
\end{figure*}
\begin{figure*}[!htbp]
  \centering
  \includegraphics[width=\textwidth]{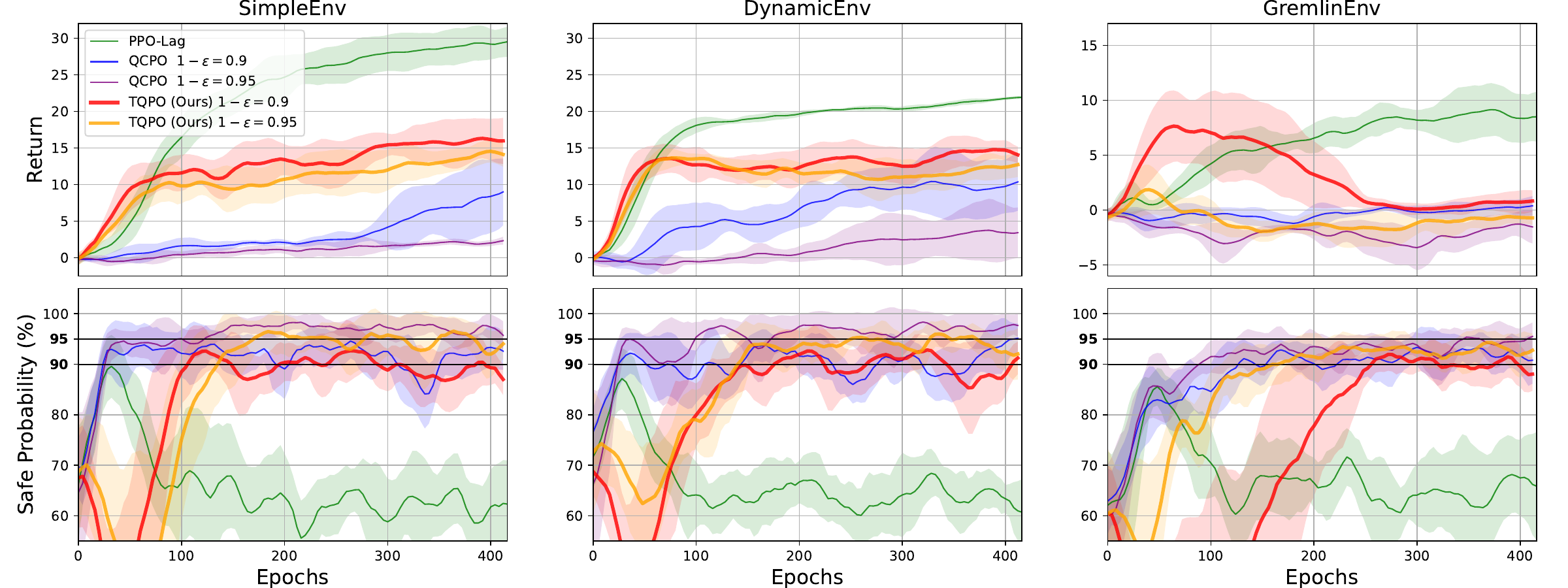}
  \caption{Return (Row 1) and Safety Probability (Row 2) of three algorithms on SimpleEnv (Column 1), DynamicEnv (Column 2) and GremlinEnv (Column 3).}
  \label{fig:results}
\end{figure*}

The reward is defined as the distance reduction from the robot to the goal over two time steps. When the robot reaches the goal, the environment provide an additional reward $+1$ and moves the goal to a fixed (Fig.~\ref{fig:simp_env}) or random position (Fig.~\ref{fig:dyna_env}, \ref{fig:gem_env}). When the robot collides with other objects, it receives a cost $+1$ at this time step, otherwise the cost is $0$. This reward and cost shaping encourages the robot to reach the goal as many times as possible while avoiding collisions in an episode of 1000 steps.

\subsection{Baseline Methods and Evaluation}

We compare the proposed TQPO with the state-of-the-art quantile-constrained RL algorithm QCPO in \cite{jung2022quantile} and a classic expectation-constrained RL method PPO-Lag in \cite{stooke2019rlpyt}. Four metrics are used to evaluate the performance of the algorithms: Average episode return, safety probability, average episode cost and $1-\varepsilon$ quantile of cost. The cost threshold is set to $d=15$. Since many safety-critical applications require a high safety probability above $90\%$, $1-\varepsilon=90\%,95\%$ are used in the experiments. All the experiments are conducted with five random seeds, with the solid line representing the mean and shaded area indicating the standard deviation. Implementation details can be found in Appendix B\footnote{Code is available at \url{https://github.com/CharlieLeeeee/TQPO}}.

\subsection{Results}

First, we prove that compared to the quantile constraint, \textbf{the expectation constraint is not suitable for safety-critical scenarios}. Fig.~\ref{fig:results_cost} shows the average cost (Row 1) and $1-\varepsilon$ quantile of cost (Row 2). From Row 1, we can observe that PPO-Lag (green) satisfies its expectation constraint $\mathbb{E}[C]\le d$, with the average cost around the threshold (black line). However, as shown in Row 2, the $90\%$ cost quantile of PPO-Lag significantly exceeds the threshold. In contrast, both QCPO (blue and purple) and TQPO (red and orange) satisfy their quantile constraints $q_{1-\varepsilon}(\pi_\theta)\le d$, with their cost quantiles around the threshold (black line) in Row 2, and their average cost is below that of PPO-Lag in Row 1. This difference in cost between the expectation-based method and the quantile-constrained methods results in the difference in safety probability. Fig.~\ref{fig:results} Row 2 demonstrates that the safety probability of PPO-Lag is below $70\%$, which is significantly lower than that of QCPO and TQPO. As discussed in the introduction, the expectation constraint fails to bound the safety probability, resulting in the low safety probability of PPO-Lag. In comparison, QCPO and TQPO achieve higher safety probabilities closer to the given level (black line) in Fig.~\ref{fig:results} Row 2. 
Therefore, in safety-critical scenarios with high safety probability requirements, the quantile constraint can achieve better safety performance and is more suitable than the expectation-based constraint.

Next, we compare the two quantile-constrained methods. For a more intuitive comparison, we evaluate the average performance of trained QCPO and TQPO algorithms, as shown in Table \ref{tab:results}. Higher return and closer safety probability to the given level are preferred, as highlighted by the bolded values.
\begin{table}[h]
    \centering
    \fontsize{9}{10}\selectfont
    \begin{tabular}{c c c c}
      \hline
      Tasks & Metrics & QCPO & TQPO \\
      \hline
      \multirow{3}{*}{SimpleEnv 90\%} & R & 8.6±0.4 & \textbf{16.1±0.2} \\
            & Pr & 92\%±1\% & \textbf{89\%±1\%} \\
            & T(h) & 3.0±0.1 & \textbf{2.7±0.1} \\
            \hline
            \multirow{3}{*}{SimpleEnv 95\%} & R & 2.1±0.2 & \textbf{14.2±0.3} \\
            & Pr & 97\%±1\% & \textbf{95\%±1\%} \\
            & T(h) & 3.2±0.2 & \textbf{2.6±0.1} \\
            \hline
            \multirow{3}{*}{DynamicEnv 90\%} & R & 10.0±0.3 & \textbf{14.4±0.4} \\
            & Pr & 91\%±1\% & \textbf{90\%±1\%} \\
            & T(h) & 2.6±0.2 & 2.6±0.1 \\
            \hline
            \multirow{3}{*}{DynamicEnv 95\%} & R & 3.5±0.1 & \textbf{12.5±0.2} \\
            & Pr & 97\%±0\% & \textbf{94\%±1\%} \\
            & T(h) & \textbf{2.7±0.1} & 3.0±0.2 \\
            \hline
            \multirow{3}{*}{GremlinEnv 90\%} & R & 0.33±0.1 & \textbf{0.77±0.1} \\
            & Pr & 92\%±1\% & \textbf{90\%±1\%} \\
            & T(h) & 4.4±0.3 & \textbf{3.4±0.1} \\
            \hline
            \multirow{3}{*}{GremlinEnv 95\%} & R & -1.43±0.2 & \textbf{-0.66±0.1} \\
            & Pr & \textbf{94\%±1\%} & 93\%±1\% \\
            & T(h) & 4.0±0.2 & \textbf{3.6±0.2} \\
      \hline
    \end{tabular}
    \caption{Empirical results of QCPO and TQPO on three tasks with safety probability $1-\varepsilon=90\%, 95\%$. R: Average episode return, Pr: Safety probability, T: Training time.}
    \label{tab:results}
\end{table}

First consider \textbf{the safety probability of QCPO and TQPO}. Fig.~\ref{fig:results} Row 2 shows that both QCPO (blue for $1-\varepsilon=90\%$, purple for $1-\varepsilon=95\%$) and TQPO (red for $1-\varepsilon=90\%$, orange for $1-\varepsilon=95\%$) achieve safety probability close to the given level $1-\varepsilon$ (black line). However, the curves of QCPO are more likely to be above $1-\varepsilon$ rather than around the given probability level like TQPO. Table \ref{tab:results} (Pr Columns) also demonstrates that the safety probability of TQPO is closer to the given level. As mentioned before, QCPO assumes the cumulative cost follows a certain distribution and uses an additive form on $\mathbb{E}[C]$ to approximate the quantile, which may lead to biased quantile estimation, resulting in higher safety probability. In contrast, TQPO avoids any distribution assumption and expectation-form approximation, directly estimates the quantile through a sampling technique. Results in Fig.~\ref{fig:results} Row 2 and Table \ref{tab:results} demonstrate the safety probability of TQPO is closer to the given level, indicating the accuracy of our quantile gradient estimation method.

\begin{figure}
  \centering
  \includegraphics[width=0.3\textwidth]{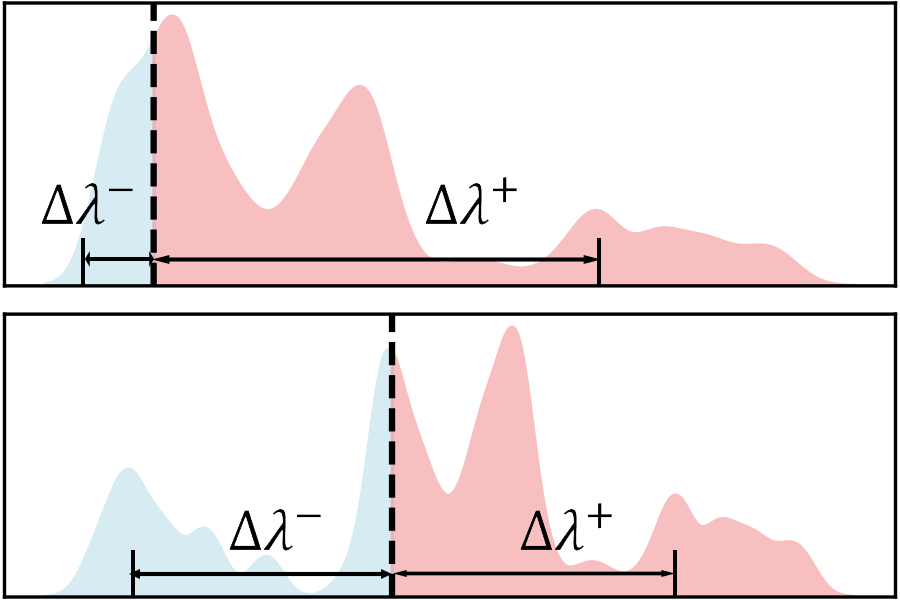}
  \caption{Distributions of quantile $q_{1-\varepsilon}$ w.o. (top) and w. (bottom) tilted term. The black vertical dashed line is the threshold $d$, $\Delta \lambda^+$ is the increase of $\lambda$ when $q_{1-\varepsilon}\ge d$, $\Delta \lambda^-$ represent the decrease of $\lambda$ when $q_{1-\varepsilon}< d$.}
  \label{fig:tilted_quantile}
\end{figure}

Next we compare \textbf{the return of QCPO and TQPO}. Fig.~\ref{fig:results} Row 1 shows that TQPO achieves higher return than QCPO in all tasks, with the return curves of TQPO above those of QCPO, especially in the case of high safety probability $1-\varepsilon=95\%$ (orange for TQPO, purple for QCPO). Table \ref{tab:results} (R Columns) also demonstrates that TQPO outperforms QCPO with higher return. Notably, in SimpleEnv and DynamicEnv (Fig.~\ref{fig:results} Row 1, Column 1\&2), TQPO with a higher safety level $1-\varepsilon=95\%$ (orange) may even outperform QCPO with a lower safety level $1-\varepsilon=90\%$ (purple). The higher return of TQPO not only proves the effectiveness of the quantile estimation method but also demonstrates the advantage of the tilted quantile gradient update. Fig.~\ref{fig:tilted_quantile} indicates that the tilted term compensates the asymmetric distributional density of the quantile, ensuring the tilted quantile symmetrically distributed around the threshold. Therefore the decrease of $\lambda$ is boosted, facilitating its recovery from overshooting and avoiding the over-conservatism of the policy, which leads to higher return for TQPO within the same number of training epochs compared to QCPO.

\begin{table}
    \centering
    \setlength{\tabcolsep}{1mm}
    \fontsize{9}{10}\selectfont
    \begin{tabular}{l cc cc}
    \hline
    \multirow{2}{*}{Algorithms}& \multicolumn{2}{c}{SimpleEnv 95\%} & \multicolumn{2}{c}{DynamicEnv 95\%}\\ 
                              & R   & Pr  & R            & Pr         \\ \hline
    QCPO                      & 2.1±0.2              & 97\%±1\%            & 3.5±0.1              & 97\%±0\%           \\
    QCPO (tilt)               & 4.7±0.2              & \textbf{95\%±1\%}           & 8.0±0.3              & \textbf{95\%±1\%}           \\
    TQPO (w/o tilt)           & 10.3±0.3             & 96\%±1\%            & 5.26±0.2             & \textbf{95\%±1\%}           \\
    TQPO (fixed tilt)         & 13.9±0.8             & 92\%±2\%            & 11.9±0.2             & 92\%±1\%           \\
    TQPO                      & \textbf{14.2±0.3}             & \textbf{95\%±1\%}            & \textbf{12.5±0.2}             & 94\%±1\%           \\ \hline
    \end{tabular}
    \caption{Ablation study on SimpleEnv 95\% and DynamicEnv 95\%. R: Average episode return, Pr: Safety probability}
    \label{tab:ablation}
\end{table}

Furthermore, Directly estimating the quantile through sampling benefits TQPO with greater time efficiency compared to QCPO, which necessitates additional time for distribution fitting and quantile approximation. As shown in Table \ref{tab:results} (T Columns), the average training time for TQPO is approximately 10\% shorter than that of QCPO. This indicates that TQPO not only surpasses QCPO in performance but also requires less time for training.

\subsection{Ablation Study}
Ablation studies are conducted on two tasks with three variants: QCPO with tilted update, TQPO w/o tilted update and TQPO with fixed tilted rates $\eta_+=0.2, \eta_-=0.8$, as shown in Table \ref{tab:ablation}. First, TQPO(w/o tilt) have higher return than QCPO, validating the effectiveness of the quantile gradient estimation. Second, QCPO(tilt) outperforms QCPO, while TQPO outperforms TQPO(w/o tilt) and TQPO(fixed tilt), indicating the benefit of proposed tilted update. The naive tilted method TQPO(fixed tilt) alleviates early overshooting of $\lambda$, but leads to its undershooting later and a biased safety probability 92\% eventually. Our tilted update calculates $\eta$ each epoch to update $\lambda$ adaptively, results in better safety probability 95\%.

\section{Conclusion}

In this paper, we have developed a novel quantile-constrained RL model named Tilted Quantile Policy Optimization. This model applies sampling-based quantile gradient estimation for quantile constraints, and a tilted quantile gradient update strategy for higher return. We provide theoretical proofs of the convergence of this TQPO model, which converges to the optimal solution under certain conditions. Experiments on three classic safe RL tasks demonstrate the effectiveness of the proposed TQPO model, which satisfies all the quantile constraints and achieves higher return than the state-of-the-art benchmarks. Our future work will focus on extending this model to multi-agent RL applications and considering the application in real-world systems.

\bigskip

\bibliography{aaai25}

\end{document}